\documentclass{article} 
\usepackage{iclr2027_conference,times}

\usepackage{amsmath,amsfonts,bm}

\def\eqref#1{equation~\ref{#1}}

\def\1{\bm{1}}

\DeclareMathAlphabet{\mathsfit}{\encodingdefault}{\sfdefault}{m}{sl}
\SetMathAlphabet{\mathsfit}{bold}{\encodingdefault}{\sfdefault}{bx}{n}

\usepackage{hyperref}
\usepackage{url}

\usepackage{color}
\usepackage{color,soul}
\usepackage{xcolor}
\usepackage[table,xcdraw]{xcolor}
\usepackage{comment}
\usepackage{amssymb}
\usepackage{bbding}
\usepackage{wrapfig}

\usepackage{multirow}
\usepackage{booktabs}
\usepackage{caption}

\usepackage{enumitem}
\usepackage{makecell}
\usepackage{longtable}
\usepackage{xtab}
\usepackage{titlesec}
\usepackage{tabularx}
\usepackage{listings}
\usepackage{tcolorbox}
\usepackage{amsmath}

\usepackage{xcolor}

\definecolor{citeorange}{RGB}{191,128,64}

\hypersetup{
    colorlinks=true,
    linkcolor=red,
    citecolor=citeorange,
    urlcolor=black
}

\definecolor{ourgreen}{HTML}{D9EDDF}
\definecolor{ourgray}{HTML}{C6C6C6}

\title{VIDEAS: Distilling Explicit Action Semantics from Demonstration Videos for World Models via Prior-Guided Simulation}

\author{%
\textbf{Jianan Wang\textsuperscript{1},
Haoquan Zhai\textsuperscript{2},
Siyang Zhang\textsuperscript{1},
Bin Li\textsuperscript{3,*},
Juan Chen\textsuperscript{1,*},
Jingtao Qi\textsuperscript{3},}\\
\textbf{Zhuo Zhang\textsuperscript{3},
Enze Wang\textsuperscript{3},
Haoxiang Jin\textsuperscript{3},
Chen Qian\textsuperscript{4}}\\
\textsuperscript{1}College of Computer Science and Technology, National University of Defense Technology \\
\textsuperscript{2}School of Electronic, Electrical and Communication Engineering, University of Chinese Academy of
Sciences \\
\textsuperscript{3}Intelligent Game and Decision Lab (IGDL) \\
\textsuperscript{4}School of Artificial Intelligence, Shanghai Jiao Tong University \\
\textsuperscript{*}Corresponding authors
}

\iclrfinalcopy
\begin{document}

\maketitle
\lhead{}

\begin{abstract}
World models learn internal representations of environment dynamics to predict future states, enabling agents to optimize action plans without physical interactions.
However, developing world models that genuinely internalize underlying causal physical laws to explicitly reason about action preconditions and subsequent state transitions remains an open challenge.
In this paper, we propose VIDEAS, a data distillation framework that transforms continuous physical dynamics from operational videos into explicit action semantics for foundation models.
Specifically, it deconstructs visual demonstrations into discrete action trajectories and utilizes advanced vision-language models (VLMs) to extract structured knowledge encapsulating action preconditions and effects.
To ensure physical consistency, we introduce a prior-guided trajectory simulation mechanism grounded within a text-based environment to rigorously validate the extracted knowledge.
Notably, we incorporate negative trajectories to enrich knowledge completeness and enhance data diversity to mitigate cognitive bias.
Furthermore, we present VIDEAS-WM, an 8B/9B-parameter suite of language-based world models trained on 34K high-quality samples derived from AgiBot-World dataset.
Extensive experiments demonstrate that VIDEAS-WM establishes state-of-the-art performance in high-level embodied action semantic reasoning, exhibiting profound physical understanding and robust generalization across unseen scenarios.
\end{abstract}

\section{Introduction}

World models enable embodied agents to simulate and refine action plans internally, circumventing the costs and risks associated with real-world trial and error \citep{lecun2022path, ha2018world, chen2025planning}. 
Currently, world models have exhibited substantial potential in facilitating tasks ranging from robotic manipulation and navigation to locomotion and policy learning \citep{dong2026learning}.
Despite these advances, learning world models with explicit knowledge of action preconditions and effects remains an open challenge \citep{hu2025text2world, xie2025making}, which entails determining the applicability of an action within the current state and predicting the subsequent environment dynamics upon execution.

Existing methods explore world models from various paradigms. 
Large language models (LLMs) are frequently leveraged as potential world models for planning \citep{hao2023reasoning} and simulation \citep{wang2025besimulator} via prompt strategies; however, their pre-training corpora capture real-world physical laws only implicitly.
Consequently, relying solely on prompting to elicit this latent knowledge often results in superficial approximations of environmental transitions and physical inconsistencies during long-horizon rollouts \citep{wang2024can}.
Observation-level generative models \citep{chen2024videocrafter2, brooks2024video} synthesize high-fidelity future states by directly reconstructing raw visual and geometric observations, but they incur significant computational overhead by predicting numerous task-irrelevant details.
Existing latent-space approaches \citep{assran2023self, assran2025v} model environmental dynamics within compressed embeddings, which inherently sacrifice interpretability and struggle to preserve fine-grained semantics.
Overall, current research often emphasizes prediction accuracy while neglecting the underlying causal dependencies governing action execution.

\begin{wrapfigure}{r}{0.52\columnwidth}
    \centering
    \includegraphics[width=\linewidth]{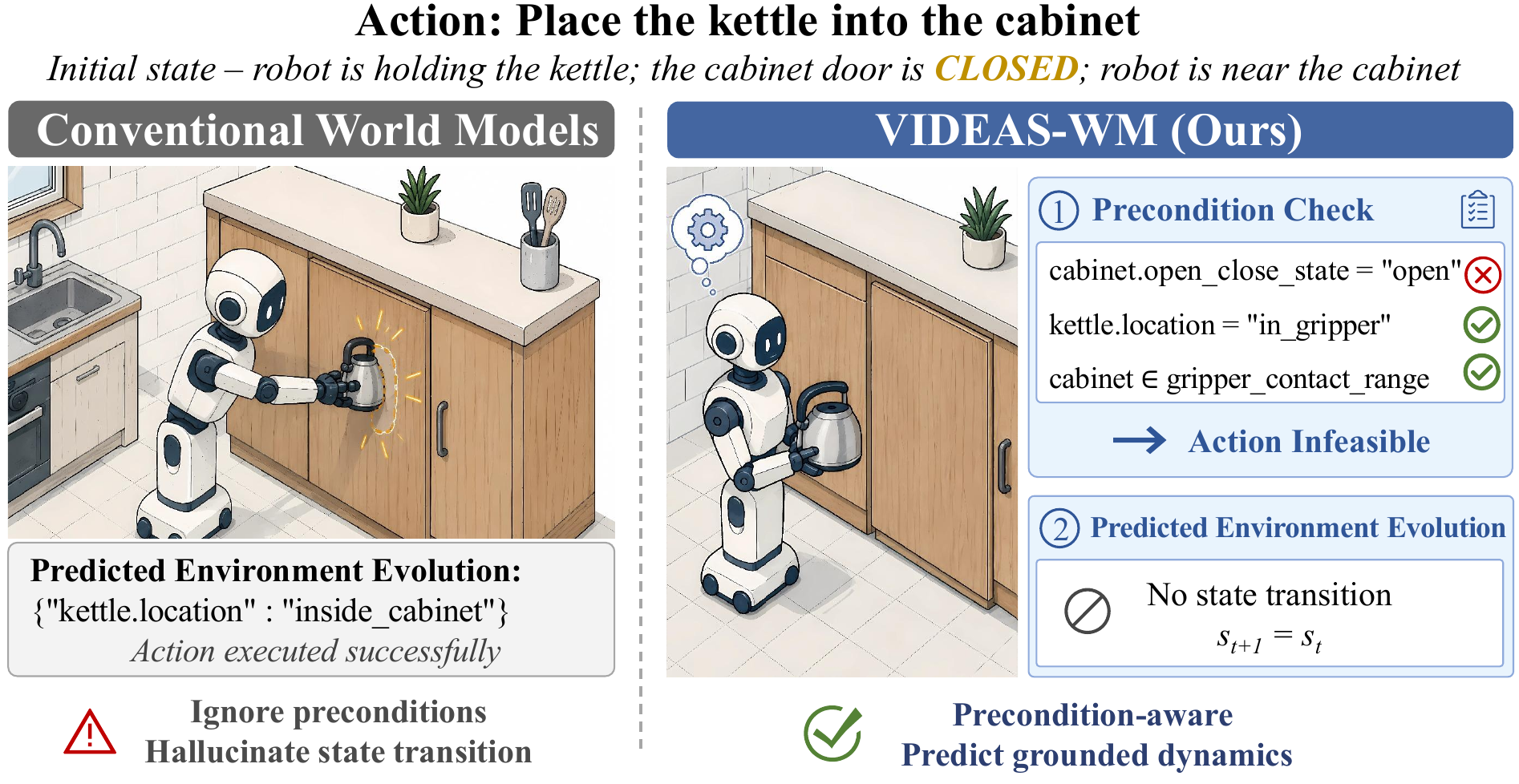}
    \caption{Conventional world models vs. VIDEAS-WM. For an infeasible action, the former ignores action preconditions and hallucinates successful state transitions, while VIDEAS-WM verifies preconditions before predicting environment evolution, correctly judging the action infeasible.}
    \label{fig:figure1}
\end{wrapfigure}

To bridge this gap, we propose \textbf{VIDEAS} (\textbf{VI}deo-\textbf{D}istilled \textbf{E}xplicit \textbf{A}ction\textbf{ S}emantics), a novel framework that distills action semantics from in-the-wild operational videos into foundation models, establishing high-level world models equipped with structured knowledge of action preconditions and effects. 
Inspired by humans developing physical understanding through observing environmental dynamics, we leverage massive online corpora of operational videos that inherently encapsulate the temporal and causal dependencies of actions and the resulting environmental evolution.
Rather than superficial video transcription, VIDEAS represents the video demonstrations as discrete action trajectories and leverages advanced VLMs to extract structured knowledge from the multimodal data.
To ensure physical correctness, we introduce a trajectory simulation mechanism guided by execution priors to rigorously validate extracted action semantics. 
This mechanism simulates the given trajectory within a text-based environment to progressively derive the semantics of individual actions, and verifies the simulated outcomes against execution priors, thereby filtering out physically inconsistent knowledge.
Furthermore, we deliberately construct failed trajectories to improve the completeness of action precondition knowledge and enhance data diversity.
As a further step, we introduce \textbf{VIDEAS-WM}, an 8B/9B-parameter suite of language-based world models trained on 34K high-quality samples curated from AgiBot-World dataset \citep{bu2025agibot}.

We conduct comprehensive experiments to evaluate the capability of VIDEAS-WM in robot behavior simulation and high-level world modeling.
Experimental results across multiple benchmarks demonstrate that VIDEAS-WM effectively internalizes embodied action semantics, attaining state-of-the-art performance.
Specifically, VIDEAS-WM achieves average performance gains of 10.67\% on BTSIMBENCH \citep{wang2025besimulator} and 7.00\% on LogicEnvEval \citep{wang2026logicenvgen} compared to the strong Gemini 3.1 Pro baseline \citep{gemini-3.1-pro}.
Furthermore, VIDEAS-WM exhibits strong cross-scenario generalization on challenging WorldPrediction tasks \citep{chen2025worldprediction}, outperforming larger open-source foundation models.

Overall, our work makes the following key contributions:
\begin{itemize}[leftmargin=*, itemsep=0pt, topsep=0pt]
    
    \item We propose VIDEAS, a scalable data distillation framework that transforms the continuous physical dynamics of operational videos into explicit action semantics for foundation models. It features a prior-guided trajectory simulation mechanism and the incorporation of failed trajectories to ensure semantic correctness and enhance physical grounding.

    \item Leveraging the VIDEAS framework, we automatically curate a high-quality corpus of 34K samples and present VIDEAS-WM, an 8B/9B-parameter suite of language-based world models.    

    \item Experimental results show that VIDEAS-WM achieves state-of-the-art performance in high-level embodied action semantic reasoning, validating its practical utility on downstream applications including robotic simulation and reasoning.
    
\end{itemize} 

\section{Related Work}

\textbf{World Models.}
Existing work explores world models through distinct paradigms.
Observation-level generative models synthesize future states by reconstructing visual and geometric observations \citep{DBLP:conf/iclr/YangDGTKSA24, Genie, 4D-fy}.
While high-fidelity, they incur heavy computational overhead and struggle to maintain physical consistency over a long horizon.
Latent-space methods \citep{assran2023self, DBLP:journals/corr/abs-2403-00504} model environmental dynamics within compressed abstract representations, which capture underlying motion and semantic structures but sacrifice interpretability.
Reinforcement learning-based approaches, exemplified by the Dreamer series \citep{DBLP:conf/iclr/HafnerLB020, DBLP:journals/nature/HafnerPBL25}, function as differentiable simulators to learn policies through imagined rollouts.
Recently, Cosmos 3 \citep{DBLP:journals/corr/abs-2606-02800} presents an open-weight family of omnimodal world models that jointly model language, image, video, audio, and action, supporting multimodal reasoning, world simulation, and action generation within a unified architecture.
Additionally, some work \citep{DBLP:journals/corr/abs-2511-02225, DBLP:journals/corr/abs-2507-03298} factorizes the environment into interacting entities to capture object-level dynamics.
Another active line \citep{hao2023reasoning, DBLP:conf/acl/WangFCZCFQ25, DBLP:journals/corr/abs-2512-18832} uses LLMs to model abstract transitions through natural language, yielding greater interpretability and computational efficiency.
However, relying on implicit physical knowledge from pre-training renders their long-horizon predictions brittle \citep{wang2024can}.
Moreover, existing methods primarily focus on future state prediction but overlook assessing action feasibility before execution—a gap that our work is designed to address.

\textbf{Learning from Demonstration Videos.}
Demonstration videos naturally encapsulate physical interactions and causal dynamics, serving as a scalable resource for robotics learning.
One prominent direction \citep{DBLP:conf/cvpr/ChenSZPL25, DBLP:conf/corl/LiZX0SPZ24, DBLP:journals/corr/abs-2409-16283} focuses on learning low-level visuomotor policies from video data.
For instance, LAPA \citep{DBLP:conf/iclr/YeJJJYPMTCLLL0Z25} infers discrete latent actions via unsupervised pre-training, bridging actionless internet video and policy learning.
Parallel efforts \citep{DBLP:conf/iros/WangZDFF25, DBLP:conf/iros/XieWXWC25} transform operational videos into code-based policies for high-level task planning.
To capture explicit causal logic, some work \citep{DBLP:journals/corr/abs-2505-18382, DBLP:journals/corr/abs-2507-21545} derives formal symbolic domains from demonstrations, which inherently restricts their open-world scalability.
VLWM \citep{chen2025planning} learns world models on natural videos by using natural language as a descriptive abstraction to represent action-conditioned state transitions.
In contrast, we distill real-world demonstration videos into structured action semantics, explicitly formulating action preconditions and effects to equip world models with grounded physical logic.

\section{Method}
\label{sec:method}

\subsection{Problem Formulation}
In this work, we learn a high-level embodied world model $f_{\theta}$ to internalize explicit action semantics. 
Given the current environment state $s_{t}$ and an intended robot action $a_t$, the model deduces a set of requisite preconditions $p_t$ for the action, and predicts the state transitions $\Delta s_{t}$ upon execution:
\begin{equation}
    \begin{gathered}
        p_t \sim f_\theta\big(\cdot \mid \text{config}_{\text{pre}}, s_t, a_t\big), \\
        \Delta s_t \sim f_\theta\big(\cdot \mid \text{config}_{\text{eff}}, s_t, a_t\big),
    \end{gathered}
    \label{eq:dynamics}
\end{equation}
where $\text{config}$ represents the system prompt, $s_t$ denotes the text-based environment observation, $a_t$ denotes the textual action instruction, and $p_t$ denotes the set of textual action preconditions. 
The subsequent state is then given by $s_{t+1} = s_{t} \oplus \Delta s_{t}$, where $\oplus$ applies the transitions to $s_t$.

Furthermore, to determine the action executability, the model must ground these preconditions in the concrete environmental context.
This process entails assessing whether an individual condition $c\in p_t$ is satisfied under the current state $s_{t}$:
\begin{equation}
    v \sim f_\theta\!\left(\cdot \mid \text{config}_{\text{grd}}, s_t, c\right),
    \label{eq:grounding}
\end{equation}
where $v \in \{\text{True}, \text{False}\}$ is a boolean indicator reflecting the fulfillment of the condition.

\begin{figure*}[htbp]
    \centering
    \includegraphics[width=\linewidth]{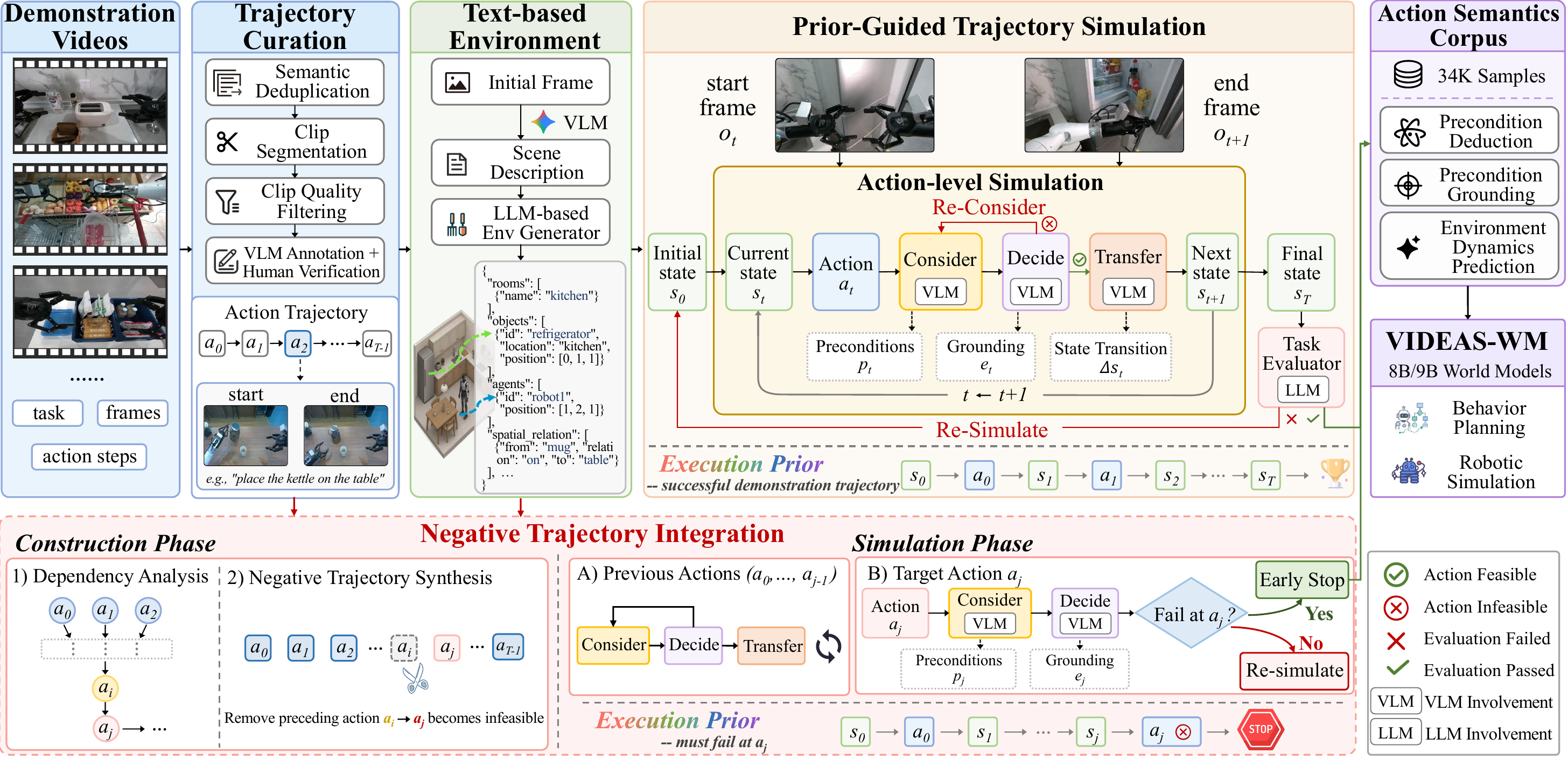}
    \caption{Overview of VIDEAS. 
    Based on in-the-wild demonstration videos, VIDEAS first curates them into discrete action trajectories and grounds them into text-based structured environments, where it leverages VLMs to iteratively simulate each action via \textit{Consider–Decide–Transfer} phases under execution priors. 
    Moreover, it constructs negative trajectories via dependency-aware mutation to enrich precondition knowledge. 
    Finally, the distilled 34K-sample corpus serves to train VIDEAS-WM, an 8B/9B-parameter suite of language-based world models.}
    \label{fig:overview}
    \vspace{-12pt}
\end{figure*}


\subsection{Data Preprocessing}
Learning a physically-grounded embodied world model requires a high-quality corpus of action semantics that encompasses causal knowledge of action preconditions and effects.
As illustrated in Figure \ref{fig:overview}, we present our data preprocessing pipeline based on the AgiBot-World dataset \citep{bu2025agibot}, which comprises 168K long-horizon manipulation episodes accompanied by language annotations for both the overall tasks and individual action steps.

Given the abundance of semantically duplicated episodes in the raw dataset, we initially filter it to isolate a core set of unique manipulation trajectories.
Leveraging the original frame-level demarcations provided for each action step, we segment the continuous demonstrations into video clips with consistent motion and single semantic behavior.
However, these clips are highly heterogeneous in visual presentation and annotation quality.
Specifically, the raw language annotations often exhibit incomplete descriptions and contextual inconsistencies across action steps.
Thus, we employ FlowNet \citep{DBLP:conf/iccv/DosovitskiyFIHH15} to prune clips with indistinct motion, and utilize VLMs to reformulate precise action descriptions for valid clips.
We further incorporate human-in-the-loop verification to enforce semantic fidelity, guaranteeing alignment between the textual representations and visual dynamics.
Through this process, the raw demonstrations are transformed into discrete, logically coherent action trajectories $\tau = (a_0, a_1, \dots, a_{T-1})$, where $T$ is the trajectory length.


\subsection{Action Semantics Distillation}
To distill continuous physical dynamics into structured action semantics, we introduce a prior-guided trajectory simulation mechanism established within text-based environments.
The mechanism progressively dissects individual embodied actions and grounds the visual demonstrations in a deterministic textual context, yielding a fully observable state representation.
This enables the precise tracking of intermediate state transitions and supports the explicit translation of action preconditions into checkable textual constraints.
Furthermore, language inherently abstracts away the low-level details of complex physical interactions, allowing the model to focus on the high-level causal logic that governs action preconditions and environmental transitions.


\textbf{Text-Based Environment Generation.}
For each action trajectory, VIDEAS constructs a structured, text-based simulated environment.
Specifically, it employs VLMs to extract a comprehensive scene description from the initial video frame, and then utilizes LogicEnvGen \citep{wang2026logicenvgen} to generate a JSON-formatted environment from the description.
This environment specifies the foundational configurations, such as room layouts, doors, and windows.
Within the environment, the physical entities, including robotic agents and manipulated objects, are characterized by fine-grained properties, encompassing text descriptions, 3D sizes, positions, and entity-specific attributes (e.g., \textit{open\_close\_state, indication\_reading}).
To fully contextualize the physical world, the generated environment further captures the relations among entities, including spatial relations (e.g., \textit{on, in}) and non-spatial relations (e.g., \textit{hold, grasp}).


\textbf{Prior-Guided Trajectory Simulation.}
Building upon the initial simulated environment, VIDEAS leverages advanced VLMs to simulate the action trajectory $\tau$ within a textual space, thereby incrementally distilling explicit action semantics.
Following BeSimulator \citep{wang2025besimulator}, the execution of each atomic action $a_t$ is modeled from the current state $s_{t}$ through a fine-grained three-phase process: \textit{consider}, \textit{decide}, and \textit{transfer}.
To prevent textual hallucinations and ensure physical consistency, the simulation is grounded by the visual observations before and after the action execution (i.e., the start frame $o_{t}$ and the end frame $o_{t+1}$), providing reliable visual evidence of the execution context.
During the first two phases, VIDEAS evaluates whether the action execution aligns with real-world physical logic  .
Specifically, in the \textbf{consider} phase, it extracts the necessary preconditions for successful execution:
\begin{equation}
    p_t = \mathcal{M}^{\text{con}}(s_{t}, a_t, o_{t}, o_{t+1}),
\end{equation}
where $\mathcal{M}$ denotes the teacher VLM invoked via greedy decoding (temperature = 0) during distillation, distinct from the target world model $f_\theta$.
Subsequently, in the \textbf{decide} phase, VIDEAS assesses the satisfaction of each precondition $c \in p_t$ against the textual world state and aggregates the results into an executability judgment:
\begin{equation}
    e_t = \bigwedge_{c \in p_t} v_c, \quad v_c = \mathcal{M}^{\text{dec}}(s_{t}, c),
\end{equation}
where $v_c \in \{\text{True}, \text{False}\}$ denotes the predicted satisfaction of condition $c$, and $e_t = \text{True}$ indicates that $a_t$ is feasible under $s_{t}$.
Upon validating the feasibility, VIDEAS models the environment evolution in the \textbf{transfer} phase:
\begin{equation}
    \Delta s_t = \mathcal{M}^{\text{tra}}(s_{t}, a_t, o_{t}, o_{t+1}),
\end{equation}
where $\Delta s_t$ captures the state transitions induced by the action.

However, directly relying on the unconstrained inferences of VLMs sometimes yields physically inconsistent or incomplete action semantics.
To address this, we utilize the execution priors of the action trajectories, which originate from successful real-world demonstrations.
Since these trajectories naturally guarantee the successful execution of all actions and the task completion, we use this prior to rigorously validate the extracted action semantics.
Specifically, VIDEAS uses a dual-grained iterative verification mechanism at both the action and trajectory levels.
At the action level, if an action is deemed infeasible (i.e., $e_t = \text{False}$), VIDEAS first postulates that this anomaly stems from misidentified preconditions, and then activates a \textbf{re-consider} phase to guide VLMs in rectifying the precondition knowledge.
If the action remains unexecutable after VLM reflection, which indicates incorrect state transitions in the preceding actions, VIDEAS restarts the simulation.
At the trajectory level, after the entire trajectory simulation, VIDEAS leverages LLMs to evaluate the task completion based on the final state $s_T$.
If completed, which indicates that the simulation is consistent with the prior, VIDEAS collects the action semantics distilled from the trajectory.
Conversely, completion failure reveals flawed state transitions during simulation, leading VIDEAS to conduct diagnostics and integrate the feedback into the next simulation iteration.
This process repeats until a successful simulation or a predefined iteration limit.


\textbf{Negative Trajectory Integration.}
Learning exclusively from successful demonstrations inherently restricts the world model to observing satisfied preconditions.
This hinders the model from generalizing the learned causal knowledge to evaluate action feasibility when physical preconditions are unmet.
Therefore, we introduce negative action trajectories, which fail at the predetermined step due to unfulfilled preconditions.
This integration enriches the semantic completeness of precondition knowledge and provides diverse negative samples for action precondition assessment.

To construct negative trajectories, we mutate the successful action trajectories based on the action execution dependencies.
Specifically, we first define a set of common dependency types $\mathcal{D}$, including possession dependencies (e.g., \textit{requiring a previously grasped object}) and state dependencies (e.g., \textit{relying on a previously opened cabinet}).
We then employ advanced LLMs to analyze the causal relationships among actions.
This yields a dependency record associating each action $a_k$ with the preceding actions it depends on:
\begin{equation}
    \mathcal{P}_k = \{\, a_i \in \tau \mid i < k \ \text{and}\ \exists\, d \in \mathcal{D} \ \text{s.t.}\ a_i \xrightarrow{d} a_k \,\},
\end{equation}
where $a_i \xrightarrow{d} a_k$ denotes that the execution of $a_k$ depends on $a_i$ through a dependency of type $d$.
We then automatically generate a negative trajectory $\tau^{-}$ by removing a preceding action $a_i$ from the trajectory $\tau$, rendering the execution of the target action $a_j$ with $j = \min\{\, k \mid a_i \in \mathcal{P}_k \,\}$ infeasible.

For a negative trajectory $\tau^{-}$, VIDEAS simulates it under the execution prior that it must fail exactly at the predetermined action $a_j$.
In detail, VIDEAS simulates the preceding action sequence normally until reaching $a_j$, where it performs only the consider and decide phases, relying purely on the textual state with visually mismatched frames masked.
If $a_j$ is correctly judged as unexecutable, which indicates that the VLM extracts the requisite preconditions and accurately identifies their non-fulfillment within the current environment, VIDEAS triggers an early stop and collects the derived knowledge as negative samples.
In contrast, an erroneous evaluation of $a_j$ as executable violates the negative prior and forces VIDEAS to restart the simulation.


\subsection{Model Training}
We frame the training stage as optimizing the foundation model $f_\theta$ to internalize the structured action semantics.
Formally, we construct a consolidated training dataset $\mathcal{C} = \mathcal{C}_{\text{pre}} \cup \mathcal{C}_{\text{grd}} \cup \mathcal{C}_{\text{eff}}$ spanning three learning tasks, namely precondition deduction, precondition grounding, and environment dynamics prediction, whose input-output pairs are instantiated as
\begin{equation}
    \label{eq:train_pairs}
    \begin{gathered}
        \mathcal{C}_{\text{pre}} : \big((\text{config}_{\text{pre}}, s_{t}, a_t),\, p_t\big), \\
        \mathcal{C}_{\text{grd}} : \big((\text{config}_{\text{grd}}, s_{t}, c),\, v_c\big), \\
        \mathcal{C}_{\text{eff}} : \big((\text{config}_{\text{eff}}, s_{t}, a_t),\, \Delta s_t\big).
    \end{gathered}
\end{equation}
We unify these tasks into a standardized sequence generation paradigm and optimize $\theta$ with the autoregressive objective:
\begin{equation}
    \label{eq:sft_loss}
    \mathcal{L}(\theta) := - \mathbb{E}_{(x, y) \sim \mathcal{C}} \sum_{l=1}^{|y|} \log P(y_l \mid x, y_{<l}; \theta).
\end{equation}

\section{Experiments}


\subsection{Experimental Setup}


\begin{wrapfigure}{r}{0.6\columnwidth}
    \centering
    \includegraphics[width=\linewidth]{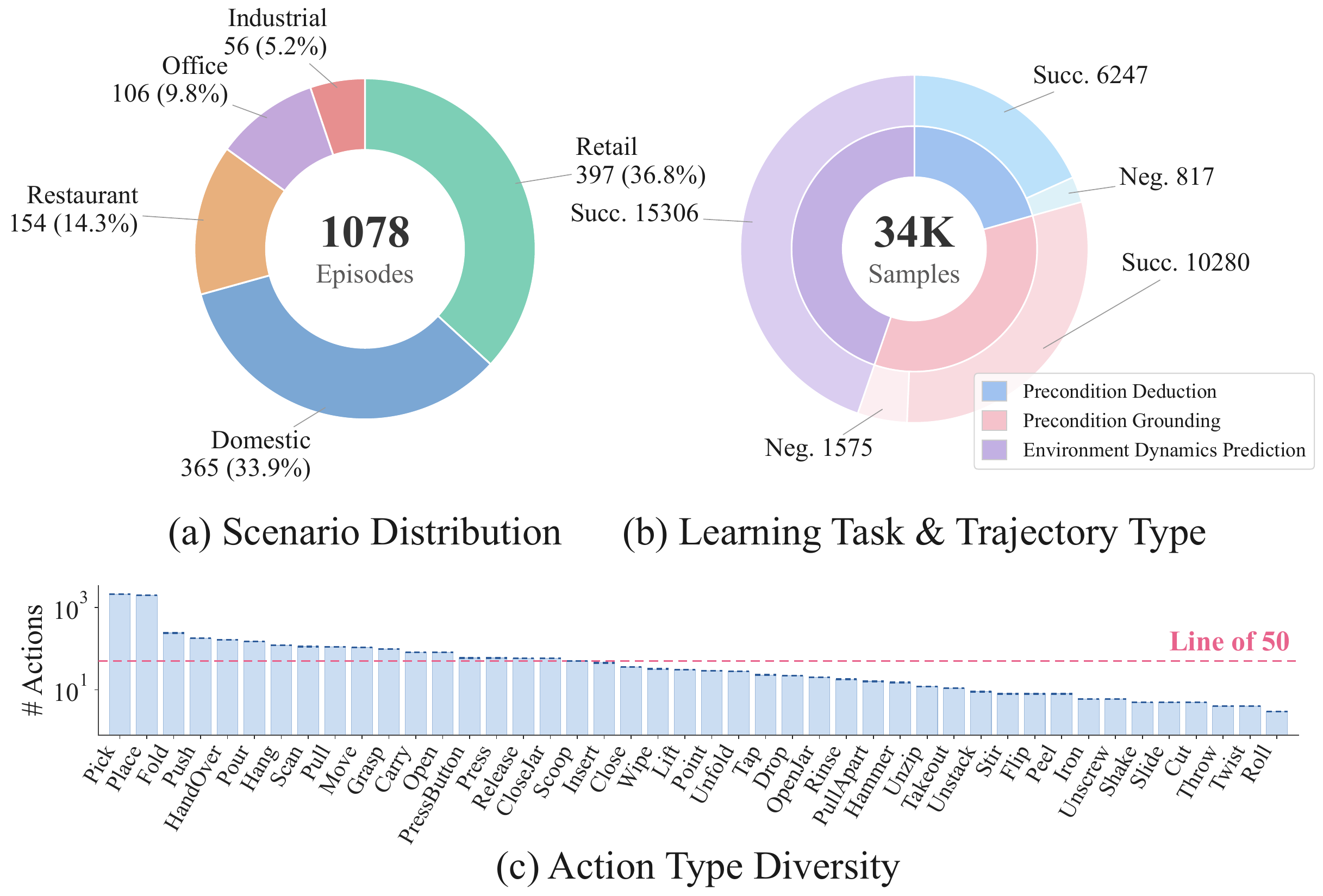}
    \caption{Training corpus statistics. In (b), Succ. and Neg. denote samples derived from successful and negative trajectories respectively. (c) shows representative action types.}
    \label{fig:data_sta}
\end{wrapfigure}

\textbf{Dataset Statistics.}
In our experiments, we utilize the AgiBot-World dataset \citep{bu2025agibot} as the demonstration data source, which contains extensive long-horizon manipulation episodes spanning diverse daily scenarios (e.g., \textit{domestic, retail, and industrial environments}).
Through rigorous deduplication and quality filtering, we extract a core subset of 1078 unique episodes covering 215 tasks and 86 action types, as illustrated in Figure \ref{fig:data_sta}.
Ultimately, we construct a high-quality corpus comprising 34K data samples, which capture explicit action semantics.


\textbf{Implementation Details.}
\label{sec:Implementation_Details}
We employ Gemini 3.1 Pro \citep{gemini-3.1-pro} as the teacher model $\mathcal{M}$, and adopt Llama-3.1-8B \citep{grattafiori2024llama}, Qwen3-8B \citep{qwen3technicalreport}, and Qwen3.5-9B \citep{qwen3.5} as the base models to instantiate VIDEAS-WM.
We perform LoRA fine-tuning for 3 epochs and the training process is conducted on 4 NVIDIA A800 80GB PCIe GPUs.
Detailed training settings are provided in the Appendix \ref{appendix:Training_Details}.


\textbf{Baselines and Benchmarks.}
To comprehensively evaluate the performance of VIDEAS-WM, we compare it against leading closed-source and open-source models across three benchmarks. Detailed simulation protocols are provided in the Appendix \ref{appendix:Benchmark_Details}.

\begin{itemize}[leftmargin=*, itemsep=0pt, topsep=0pt]
    \item \textbf{BTSIMBENCH} \citep{wang2025besimulator}: It measures how well a model can capture execution preconditions and predict action outcomes through robotic policy simulation.
    It comprises 75 robotic policies (i.e., behavior trees) categorized into Good, Counterfactuals, and Unreachable.
    We adopt the delivery rate and the simulation accuracy across these policy types as the evaluation metrics.
    The delivery rate evaluates whether the model successfully delivers the final simulation result within a predefined iteration limit.
    
    \item \textbf{LogicEnvEval} \citep{wang2026logicenvgen}: It challenges the model to determine action executability and reason about environment dynamics across long-horizon household tasks.
    Similarly, we conduct evaluations through robot behavior simulation on 100 agent policies from this benchmark.
    These policies span four categories, including the aforementioned three types and the LackBranch type.
    
    \item \textbf{WorldPrediction-WM} \citep{chen2025worldprediction}: It assesses whether the model comprehends the causalities of semantically and temporally high-level actions in daily human activities through question answering.
    Due to original data availability, we utilize a subset of 637 accessible task samples sourced from the COIN \citep{DBLP:conf/cvpr/TangDRZZZL019}, CrossTask \citep{DBLP:conf/cvpr/ZhukovACFLS19}, EPIC-KITCHENS-100 \citep{DBLP:journals/ijcv/DamenDFFKMMMPPW22}, and IKEA-ASM datasets \citep{DBLP:conf/wacv/Ben-Shabat0SCOL21}.
\end{itemize}


\begin{table*}[t]
    \centering
    \caption{Performance (\%) on BTSIMBENCH and LogicEnvEval benchmarks. Each policy category contains 25 samples. Gray and green rows represent closed-source baselines and VIDEAS-WM, respectively. The best performance among open-source models is in \textbf{bold}.}     
    \label{tab:main_results}
    
        \resizebox{\textwidth}{!}{
        \begin{tabular}{l | ccccc | cccccc}
        
        \toprule
        
            \multirow{2}{*}{\textbf{Model}} & \multicolumn{5}{c|}{\textbf{BTSIMBENCH}} & \multicolumn{6}{c}{\textbf{LogicEnvEval}} \\
            \cmidrule(lr){2-6} \cmidrule(lr){7-12}
            & Delivery & Good & CFactuals & Unreach & Avg & Delivery & Good & CFactuals & Unreach & LBranch & Avg \\
        
        \midrule
            
            \multicolumn{12}{l}{\textbf{Closed-Source Models}} \\
        
        \midrule
            
            \rowcolor{ourgray} Gemini-3.1-Pro    & 100.00 & 92.00 & 64.00 & 96.00 & 84.00 & 100.00 & 92.00 & 72.00 & 92.00 & 96.00 & 88.00 \\
            \rowcolor{ourgray} Gemini-3-Flash    & 100.00 & 88.00 & 64.00 & 88.00 & 80.00 & 100.00 & 76.00 & 76.00 & 88.00 & 96.00 & 84.00 \\
            
        \midrule
        
            \multicolumn{12}{l}{\textbf{Open-Source Models}} \\
            
        \midrule
        
            DeepSeek-V4-Pro     & 100.00 & 44.00 & 88.00 & 76.00 & 69.33 & 100.00 & 52.00 & 84.00 & 88.00 & 96.00 & 80.00 \\
            DeepSeek-V4-Flash   & 98.67 & 44.00 & 80.00 & 72.00 & 65.33 & 100.00 & 60.00 & 76.00 & 84.00 & \textbf{100.00} & 80.00 \\
            Gemma4-31B         & 100.00 & 76.00 & 72.00 & 92.00 & 80.00 & 100.00 & 88.00 & 84.00 & \textbf{92.00} & 88.00 & 88.00 \\
            Llama3.3-70B       & 98.67 & 20.00 & 92.00 & 28.00 & 46.67 & 99.00 & 32.00 & 76.00 & 48.00 & 92.00 & 62.00 \\
            Qwen-3.5-27B        & 98.67 & 40.00 & 88.00 & 76.00 & 68.00 & 100.00 & 72.00 & 80.00 & 88.00 & \textbf{100.00} & 85.00 \\
            Qwen3-32B          & 98.67 & 48.00 & 72.00 & 80.00 & 66.67 & 76.00 & 12.00 & 64.00 & 48.00 & 84.00 & 52.00 \\
            Qwen3-14B          & 98.67 & 28.00 & 84.00 & 52.00 & 54.67 & 97.00 & 32.00 & 72.00 & 60.00 & 96.00 & 65.00 \\
            Qwen2.5-72B        & 100.00 & 28.00 & 84.00 & 68.00 & 60.00 & 96.00 & 40.00 & 68.00 & 64.00 & 92.00 & 66.00 \\
            
        \midrule
        
            Llama3.1-8B        & 78.67 & 0.00 & 76.00 & 8.00 & 28.00 & 88.00 & 4.00 & 88.00 & 16.00 & 84.00 & 48.00 \\
            \rowcolor{ourgreen} VIDEAS-WM-L8B & 100.00 & \textbf{92.00} & \textbf{96.00} & \textbf{96.00} & \textbf{94.67} & 100.00 & 92.00 & 96.00 & 88.00 & 84.00 & 90.00 \\
            
        \midrule
            
            Qwen3-8B           & 85.33 & 12.00 & 80.00 & 32.00 & 41.33 & 95.00 & 16.00 & 72.00 & 40.00 & 96.00 & 56.00 \\
            \rowcolor{ourgreen} VIDEAS-WM-Q8B & 100.00 & 76.00 & \textbf{96.00} & 88.00 & 86.67 & 97.00 & \textbf{96.00} & 96.00 & \textbf{92.00} & 84.00 & 92.00 \\
            
        \midrule
        
            Qwen3.5-9B         & 86.67 & 32.00 & 72.00 & 52.00 & 52.00 & 87.00 & 28.00 & 64.00 & 64.00 & 96.00 & 63.00 \\
            \rowcolor{ourgreen} VIDEAS-WM-9B & 100.00 & \textbf{92.00} & \textbf{96.00} & \textbf{96.00} & \textbf{94.67} & 100.00 & 88.00 & \textbf{100.00} & \textbf{92.00} & \textbf{100.00} & \textbf{95.00} \\
            
        \bottomrule
        \end{tabular}
        }
    \vspace{-12pt}
\end{table*}


\subsection{Model Performance}
Table \ref{tab:main_results} presents the performance comparison between VIDEAS-WM and baselines on BTSIMBENCH and LogicEnvEval benchmarks.
The results show that VIDEAS-WM establishes state-of-the-art performance among open-source solutions, effectively assimilating explicit action semantics to achieve a deeper understanding of high-level physical logic.

Notably, our VIDEAS-WM-9B achieves average scores of 94.67\% on BTSIMBENCH and 95.00\% on LogicEnvEval, outperforming the strongest open-source model Gemma4-31B by 14.67\% and 7.00\%, respectively.
Moreover, the three VIDEAS-WM models maintain a substantial lead over their corresponding base models across most evaluation metrics, especially regarding the simulation accuracy on Good and Unreachable policies.
For instance, on the LogicEnvEval benchmark, our VIDEAS-WM-L8B achieves 92.00\% accuracy in simulating and evaluating Good policies, while the Llama3.1-8B yields 4.00\% accuracy.
This highlights the superiority of VIDEAS-WM in effectively modeling environment dynamics.
Furthermore, the performance improvement on the Counterfactuals category demonstrates the efficacy of our models in reasoning about execution preconditions and determining action executability, thereby facilitating the identification of faulty agent policies.
These results validate the practical utility of VIDEAS-WM for high-level robotic behavior simulation.


\begin{table}[t]
    \caption{Performance (\%) on WorldPrediction-WM benchmark. Green rows represent our VIDEAS-WM. The best performance within each category (VLMs and Socratic LLMs) is in \textbf{bold}.}
    \label{tab:wm_results}
    \centering

    \setlength{\tabcolsep}{5.0pt}

    \resizebox{0.85\linewidth}{!}{%
    \begin{tabular}{c l | ccccc}

    \toprule

        & \multirow{2}{*}{\textbf{Model}}
        & \multicolumn{5}{c}{\textbf{WorldPrediction-WM}} \\
        \cmidrule(lr){3-7}
        & & COIN & CrossTask & IKEA-ASM & EPIC-KITCHENS-100 & Overall \\

    \midrule

        \multirow{8}{*}{\textbf{VLMs}}
        & InternVL3.5-38B
        & 47.69 & 55.56 & 40.88 & \textbf{73.44} & 54.87 \\

        & InternVL3.5-14B
        & 47.18 & 45.56 & 35.85 & 51.56 & 45.44 \\

        & InternVL3.5-8B
        & 45.13 & 44.44 & 32.08 & 40.63 & 40.41 \\

        & Qwen3.5-27B
        & 51.28 & 55.56 & \textbf{47.80} & 67.19 & 55.82 \\

        & Qwen3-VL-32B
        & \textbf{53.85} & 54.44 & 45.28 & 69.27 & 56.45 \\

        & Qwen3-VL-8B
        & 51.79 & 55.56 & 43.40 & 67.71 & 55.03 \\

        \cmidrule(lr){2-7}

        & Qwen3.5-9B
        & 45.13 & 56.67 & 34.59 & 66.67 & 50.63 \\

        & \cellcolor{ourgreen}VIDEAS-WM-9B
        & \cellcolor{ourgreen}53.33
        & \cellcolor{ourgreen}\textbf{60.00}
        & \cellcolor{ourgreen}45.28
        & \cellcolor{ourgreen}\textbf{73.44}
        & \cellcolor{ourgreen}\textbf{58.33} \\

    \midrule

        \multirow{11}{*}{%
            \shortstack{\textbf{Socratic}\\\textbf{LLMs}}
        }
        & DeepSeek-V4-Pro
        & 63.08 & 80.00 & 32.70 & \textbf{79.27} & 62.79 \\

        & DeepSeek-V4-Flash
        & 62.56 & \textbf{81.11} & 28.30
        & \textbf{79.27} & 61.70 \\

        & MiniMax-M2.7
        & 55.90 & 71.11 & 28.93 & 68.39 & 55.10 \\

        & Llama3.3-70B
        & 51.79 & 76.67 & 10.69 & 62.69 & 48.35 \\

        & Qwen3-32B
        & 57.95 & 80.00 & 35.22 & 75.65 & 60.75 \\

        & Qwen3-14B
        & 58.97 & 74.44 & 31.45 & 74.61 & 59.03 \\

        & Qwen2.5-72B
        & 57.95 & \textbf{81.11} & 28.30 & 75.13 & 59.03 \\

        \cmidrule(lr){2-7}

        & Llama3.1-8B
        & 54.36 & 74.19 & 24.53 & 68.39 & 54.06 \\

        & \cellcolor{ourgreen}VIDEAS-WM-L8B
        & \cellcolor{ourgreen}56.92
        & \cellcolor{ourgreen}75.56
        & \cellcolor{ourgreen}38.36
        & \cellcolor{ourgreen}75.13
        & \cellcolor{ourgreen}60.43 \\

        \cmidrule(lr){2-7}

        & Qwen3-8B
        & 57.44 & 70.97 & 32.08 & 75.65 & 58.59 \\

        & \cellcolor{ourgreen}VIDEAS-WM-Q8B
        & \cellcolor{ourgreen}\textbf{65.64}
        & \cellcolor{ourgreen}76.67
        & \cellcolor{ourgreen}\textbf{41.51}
        & \cellcolor{ourgreen}73.06
        & \cellcolor{ourgreen}\textbf{63.42} \\

    \bottomrule

    \end{tabular}%
    }

    \vspace{-6mm}
\end{table}


\subsection{Cross-Scenario Generalization}
To evaluate the cross-scenario generalization capability of VIDEAS-WM, we conduct experiments on the WorldPrediction-WM benchmark.
It encompasses diverse everyday activities (e.g., \textit{household chores, furniture assembly, and nursing \& caring}).
This benchmark requires models to deduce the correct action responsible for the environment transition between the provided initial and final states.
Table \ref{tab:wm_results} shows that VIDEAS-WM-9B attains 58.33\% overall accuracy in the VLMs category, outperforming larger open-source models, such as Qwen3.5-27B and InternVL3.5-38B.
In the Socratic LLMs category, VIDEAS-WM-Q8B improves over its base model Qwen3-8B by 4.83\% and further exceeds DeepSeek-V4-Pro.
These results indicate that VIDEAS-WM abstracts universal physical logic and robustly transfers this knowledge to novel domains beyond the training distribution.


\subsection{Ablation Studies and Analysis}
We conduct ablation studies on the BTSIMBENCH benchmark to evaluate the impact of the two core components of VIDEAS, and the scaling of training data.

\begin{wraptable}{r}{0.60\textwidth}
    \vspace{-2mm}
    \centering
    \caption{Results of ablation study on the prior-guided trajectory simulation mechanism (Naive Distillation) and the negative trajectory integration (W/O NTI). }
    \label{tab:ablation_pipeline}
    \setlength{\tabcolsep}{3.5pt} 
    \renewcommand{\arraystretch}{1.1} 
    \resizebox{\linewidth}{!}{%
    \begin{tabular}{ll | cccc}
        \toprule
        \multirow{2}{*}{\textbf{Model}} & \multirow{2}{*}{\textbf{Method}} & \multicolumn{4}{c}{\textbf{BTSIMBENCH}} \\
        \cmidrule(lr){3-6}
        & & Good & CFactuals & Unreach & Avg \\
        \midrule
        
        \multirow{4}{*}{\textbf{Llama3.1-8B}} 
        & Base & 0.00 & 76.00 & 8.00 & 28.00 \\
        & Naive Distillation & 72.00 & 32.00 & 88.00 & 64.00 \\
        & W/O NTI & 88.00 & 28.00 & \textbf{100.00} & 72.00 \\
        & VIDEAS-WM & \textbf{92.00} & \textbf{96.00} & 96.00 & \textbf{94.67} \\
        
        \midrule
        
        \multirow{4}{*}{\textbf{Qwen3-8B}} 
        & Base & 12.00 & 80.00 & 32.00 & 41.33 \\
        & Naive Distillation & 52.00 & 44.00 & 76.00 & 57.33 \\
        & W/O NTI & 68.00 & 44.00 & 84.00 & 65.33 \\
        & VIDEAS-WM & \textbf{76.00} & \textbf{96.00} & \textbf{88.00} & \textbf{86.67} \\
        
        \midrule
        
        \multirow{4}{*}{\textbf{Qwen3.5-9B}} 
        & Base & 32.00 & 72.00 & 52.00 & 52.00 \\
        & Naive Distillation & 80.00 & 52.00 & 84.00 & 72.00 \\
        & W/O NTI & \textbf{92.00} & 48.00 & 92.00 & 77.33 \\
        & VIDEAS-WM & \textbf{92.00} & \textbf{96.00} & \textbf{96.00} & \textbf{94.67} \\
        
        \bottomrule
    \end{tabular}%
    }
    \vspace{-2mm}
\end{wraptable}

\textbf{Impact of Prior-Guided Trajectory Simulation Mechanism.}
To assess the efficacy of the prior-guided trajectory simulation mechanism, we construct a naive distillation baseline.
This variant directly extracts action semantics from the raw teacher annotations without the iterative verification process.
We then fine-tune the three evaluated base models on this naive dataset to observe the performance variations.
As shown in Table \ref{tab:ablation_pipeline}, the ablation variant improves the average performance over the base models by 16.00\% to 36.00\% across all three architectures.
However, our analysis reveals that the semantic knowledge distilled directly from the teacher model suffers from physical inconsistencies and biases (e.g., \textit{missing or incorrect action preconditions, and incomplete state transitions}).
Consequently, this ablation variant consistently underperforms VIDEAS-WM, exhibiting an average accuracy gap ranging from 22.67\% to 30.67\%.
This highlights the vital role of our mechanism in ensuring knowledge correctness and providing high-quality supervisory signals.

\textbf{Impact of Negative Trajectory Integration.}
To evaluate the effectiveness of incorporating negative trajectories, we construct a data subset by excluding 2.4K negative samples, which exclusively contains the distilled knowledge from successful trajectories.
Table \ref{tab:ablation_pipeline} compares models trained on this subset and the full dataset.
The results reveal a decline in accuracy across most categories for all three models, particularly in the Counterfactuals category.
For Qwen3.5-9B, the accuracy of this category drops from 96.00\% to 48.00\%, which even underperforms the base model.
Essentially, the Counterfactuals category evaluates the capability of the model to identify unmet execution preconditions.
Thus, exposure to merely positive instances induces severe cognitive bias, causing the model to overfit and blindly predict successful state transitions.
This confirms the significance of negative trajectories in enhancing data diversity and knowledge completeness, which enables the world model to develop more comprehensive physical understanding of action execution.

\begin{figure*}[t] 
    \centering
    \includegraphics[width=1.0\textwidth]{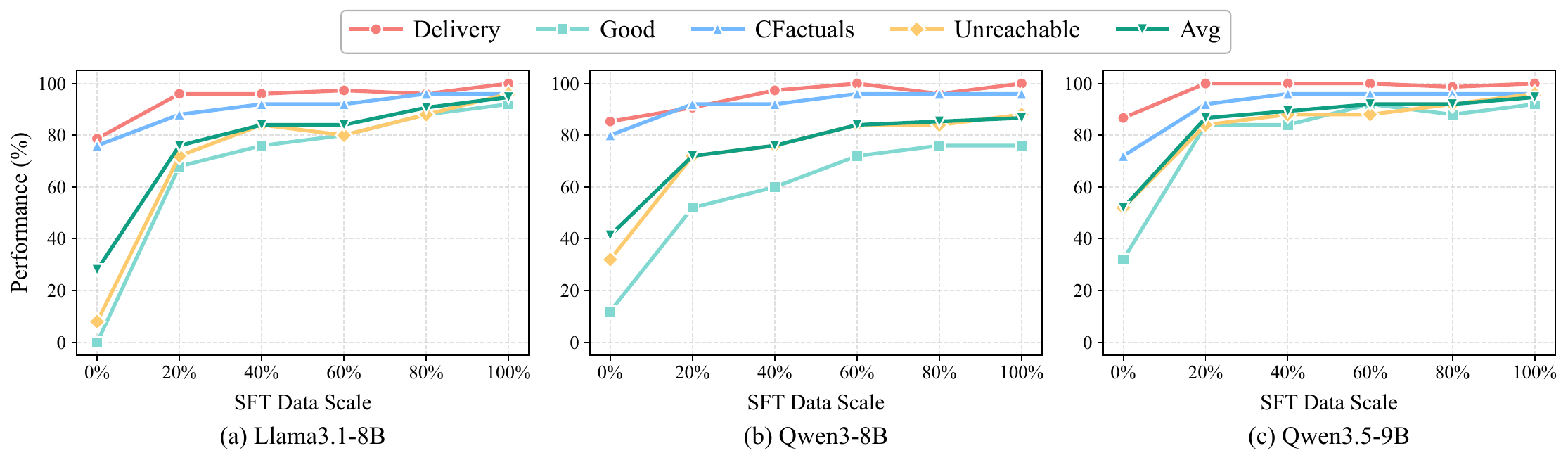}
    \caption{Ablation study results on training data scales for the BTSIMBENCH benchmark.}
    \label{fig:ablation_datascale}
    \vspace{-4mm}
\end{figure*}

\textbf{Data Scaling for Model Fine-Tuning.}
To investigate the influence of training data scale, we uniformly sample 20\%, 40\%, 60\%, and 80\% of the full training corpus to perform supervised fine-tuning (SFT) on the three base models.
To ensure consistency, we employ a fixed random seed to shuffle the dataset and extract prefixes of the same permutation, guaranteeing that smaller subsets are strict subsets of the larger ones.
As shown in Figure \ref{fig:ablation_datascale}, the performance scales consistently with the dataset size before reaching a plateau.
For instance, fine-tuning the Llama3.1-8B model with merely 20\% of the data yields a substantial performance leap, improving the average accuracy from 28.00\% to 76.00\%.
Further increasing the data scale to 100\% pushes the average accuracy to 94.67\%.
This saturation trend suggests that while our dataset effectively injects action semantics into the models, further performance gains through simply scaling the supervised data become marginal.

\textbf{Human Verification for Distilled Corpus.}
We conduct human verification of the knowledge extracted by VIDEAS.
Specifically, we randomly sample 135 action instances from the 34K corpus and have three independent annotators assess whether the distilled action preconditions and effects align with real-world physical logic. 
Final labels are determined via majority voting, and we observe a high inter-annotator agreement.
The evaluation reveals accuracies of 97.78\% for preconditions and 98.52\% for effects.
We analyze these samples and find that the few invalid precondition samples primarily suffer from missing necessary conditions.
For example, the action of placing a scouring pad onto a counter verifies whether the robot holds the pad but misses checking whether the counter is within the gripper contact range.
This observation further corroborates the necessity of negative trajectory integration in improving the completeness of precondition knowledge.
Meanwhile, the few invalid effect samples mainly result from incomplete state transition predictions.

\section{Conclusion}

We propose VIDEAS, a VLM-powered data distillation framework that transforms continuous physical dynamics from demonstration videos into explicit action semantics for world models.
To ensure the logical correctness and completeness of the extracted semantics, we introduce a prior-guided trajectory simulation mechanism and construct negative trajectories.
Furthermore, we present VIDEAS-WM, a suite of language-based world models.
Experimental results across multiple benchmarks show that VIDEAS-WM achieves superior performance in embodied action semantic reasoning, exhibiting deep physical understanding and robust generalization in novel scenarios.
Our work provides new insights for internalizing explicit physical causalities within world models to facilitate high-level planning and robotic simulation.

\section*{Ethics statement}

The development of VIDEAS-WM strictly adheres to established ethical guidelines regarding data usage and model deployment.
All raw operational videos and textual annotations originate from publicly available datasets under permissive licenses.
For human-in-the-loop verification, we ensure that all annotators receive fair compensation and participate voluntarily with informed consent.
Our research exclusively utilizes open-source foundation models to promote methodological transparency and ensure scientific reproducibility.
While VIDEAS-WM advances embodied action semantic reasoning, we explicitly restrict its intended use to academic research.
Users must be aware that the model may exhibit hallucinations and occasionally generate physically misleading outcomes.
Therefore, we urge that applications of this model undergo safety checks to prevent unpredictable hazards.

\section*{Reproducibility statement}

To ensure the reproducibility of our work, we describe the data preprocessing pipeline and the trajectory simulation mechanism in Section \ref{sec:method}.
The exact prompt templates used for this mechanism are provided in Appendix \ref{app:prompt}.
We specify the model training setups and essential hyperparameters in Section \ref{sec:Implementation_Details} and Appendix \ref{appendix:Training_Details}.
The detailed evaluation protocols for all benchmarks are available in Appendix \ref{appendix:Benchmark_Details}.
Moreover, We submit the source code of the VIDEAS framework as supplementary material to facilitate future research.

\bibliography{iclr2027_conference}
\bibliographystyle{iclr2027_conference}

\vfill
\newpage

\appendix


\section{Limitations}

First, VIDEAS establishes the trajectory simulation and semantics extraction within a text-based environment.
We adopt this formulation because natural language effectively abstracts complex visual observations and highlights the high-level causal logic of embodied actions.
Although this abstraction facilitates efficient world modeling, it inherently discretizes state representations and discards continuous physical constraints such as kinematics, collision geometry, and fluid dynamics.
Consequently, the current framework functions primarily as a semantic logic engine rather than a fully grounded continuous physics simulator.
Second, the data distillation pipeline relies on VLMs to extract action semantics.
Although we employ the prior-guided trajectory simulation mechanism to rigorously validate the extracted knowledge, latent hallucinations from VLMs might still introduce subtle noise into the knowledge corpus.


\section{Training Details}
\label{appendix:Training_Details}

For model training, we apply the Low-Rank Adaptation technique to update all linear projections within the attention mechanisms and feed-forward networks.
While Qwen3.5-9B serves as a VLM, we perform pure text fine-tuning and keep its visual encoder frozen.
The training process takes approximately 7 hours for the Llama-3.1-8B-Instruct model, 7.5 hours for the Qwen3-8B model and 16 hours for the Qwen3.5-9B model (on four NVIDIA A800 GPUs).
Table \ref{tab:training-hyperparameters} presents the essential training hyperparameter configurations in our experiments.
Additionally, the temperatures of LLMs and VLMs are set to 0 during evaluation.

\begin{table}[h]
    \centering
    \caption{Training hyperparameters.}
    \label{tab:training-hyperparameters}
    \setlength{\tabcolsep}{13pt}
    
    \begin{tabular}{lc}
        \toprule
            \textbf{Hyperparameter} & \textbf{Value} \\
        \midrule
            Training epochs & 3 \\
            Learning rate & $1 \times 10^{-4}$ \\
            Warmup ratio & 0.03 \\
            Batch size & 4 \\
            Gradient accumulation steps & 1 \\
            LoRA rank & 32 \\
            LoRA alpha & 64 \\
            LoRA dropout & 0.1 \\
            Precision & bfloat16 \\
            Maximum sequence length & 4096 \\
        \bottomrule        
    \end{tabular}
        
\end{table}


\section{Evaluation Details}
\label{appendix:Benchmark_Details}

Our evaluation benchmarks comprise BTSIMBENCH \citep{wang2025besimulator}, LogicEnvEval \citep{wang2026logicenvgen}, and WorldPrediction-WM \citep{chen2025worldprediction}.
During the evaluation on BTSIMBENCH and LogicEnvEval, we explicitly provide two few-shot examples in the system prompt for the evaluated foundation models to enhance their instruction following capabilities.

\subsection{BTSIMBENCH}

For the BTSIMBENCH evaluation, we integrate various foundation models into the BeSimulator framework \citep{wang2025besimulator}.
This integration allows us to assess the action semantic reasoning capabilities of the models by deploying them as the core predictive engines within the embodied simulated environment.
This benchmark contains three categories of robot policies, and we adopt the delivery rate and the simulation accuracy across these three categories as evaluation metrics.
Specifically, the delivery rate measures whether the model successfully delivers the final simulation result within a predefined iteration limit (5 times in our experiment settings).
Common reasons for delivery failure include response timeouts and formatting errors.
The simulation accuracy represents the ratio of policies that the model correctly evaluates for each specific category.
It requires the model to precisely capture execution preconditions and predict environment transitions during the simulation process.

\subsection{LogicEnvEval}

The LogicEnvEval benchmark is originally used to evaluate simulated environment generation.
We utilize its task data and 100 agent policies spanning four distinct categories.
Then, we evaluate the foundation models through robot behavior simulation like BTSIMBENCH.

\subsection{WorldPrediction-WM}

Regarding the WorldPrediction-WM benchmark, we encounter accessibility and permission constraints with several original source video datasets (COIN \citep{DBLP:conf/cvpr/TangDRZZZL019}, CrossTask \citep{DBLP:conf/cvpr/ZhukovACFLS19}, and EgoExo4D \citep{DBLP:conf/cvpr/GraumanWTKMAABB24}).
Consequently, we use an evaluation subset consisting of 637 accessible task samples.
To ensure fair comparisons, we evaluate all foundation models exclusively on this exact subset.
Table \ref{tab:worldprediction-stats} details the precise distribution of these samples.
Given the initial and final world states, this benchmark requires the evaluated models to deduce the correct action responsible for the environment transition from a set of counterfactual distractors.   
For VLMs, these states are directly provided as visual inputs comprising the start and end video frames.
Because VIDEAS-WM-9B retains the frozen visual encoder from its base model, it directly processes the visual frames and is thereby categorized under the VLMs group during evaluation.
For the Socratic LLMs such as DeepSeek-V4-Pro, we follow the official setting of WorldPrediction-WM, which decouples perception and reasoning into two separate stages.
Specifically, the visual inputs are first translated into textual descriptions through a VLM, and the LLM under evaluation is then prompted with these captions together with the structured task explanations and candidate actions.
The LLM subsequently performs textual reasoning to identify the correct action.
To obtain the textual descriptions, we follow the official setting and utilize Qwen2.5-VL-72B \citep{DBLP:journals/corr/abs-2502-13923}.

\begin{table}[h]
    \centering
    \caption{Distribution of accessible task samples in WorldPrediction-WM.}
    \label{tab:worldprediction-stats}
    \setlength{\tabcolsep}{7pt}
    
    \begin{tabular}{lc}
        \toprule
            \textbf{Dataset} & \textbf{Number of Samples} \\
        \midrule
            COIN & 195 \\
            CrossTask & 90 \\
            IKEA-ASM & 159 \\
            EPIC-KITCHENS-100 & 193 \\
        \midrule
            Total & 637 \\
        \bottomrule
    \end{tabular}

\end{table}


\section{Efficiency Analysis}

\begin{figure}[h]
    \centering
    \includegraphics[width=0.75\linewidth]{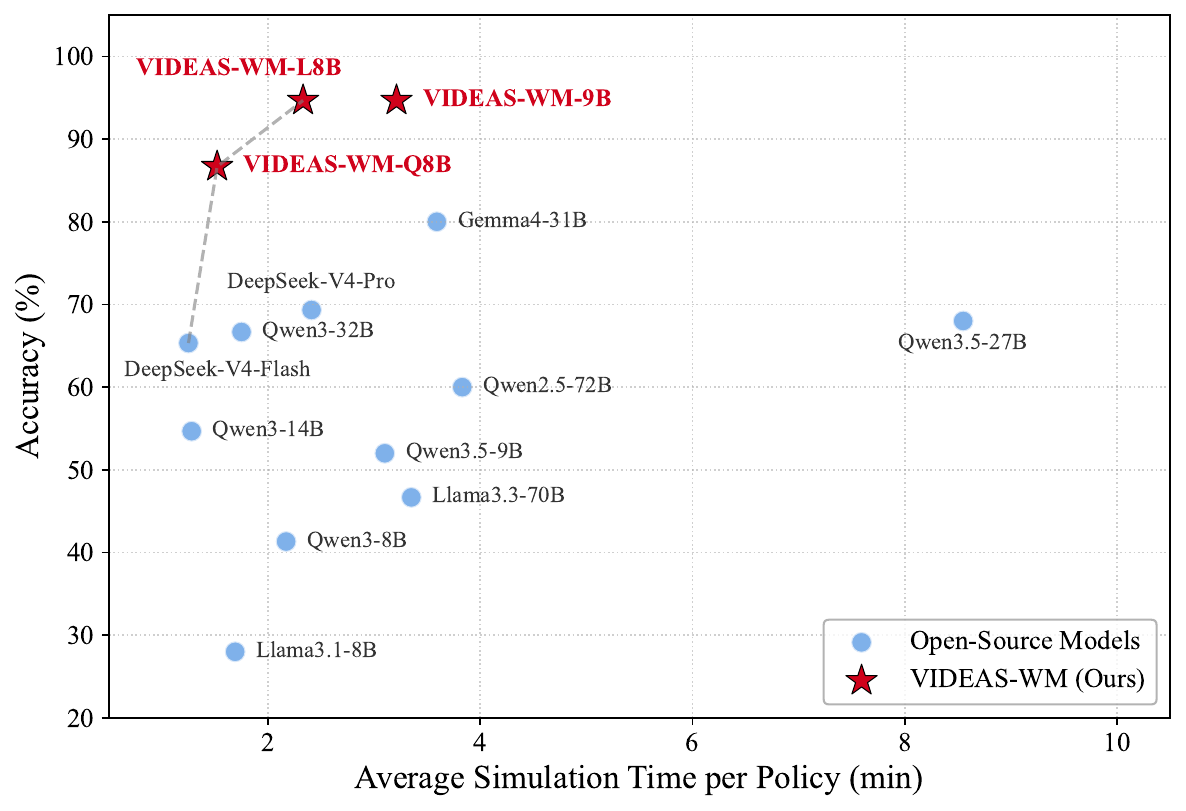}
    \caption{Pareto frontier of accuracy versus average simulation time per robot policy. VIDEAS-WM establishes the optimal balance between reasoning capability and computational efficiency.}
    \label{fig:efficiency}
\end{figure}

We conduct an efficiency analysis on the BTSIMBENCH benchmark to evaluate the inference latency of various foundation models.
Specifically, we measure the average time required to simulate a single robot policy in minutes.
To intuitively illustrate the performance and efficiency trade-off, we present a Pareto frontier scatter plot in Figure \ref{fig:efficiency}, where the horizontal axis denotes the simulation time cost and the vertical axis represents the average simulation accuracy.
Models situated in the top-left region of this plot achieve the optimal balance between high reasoning capability and low computational overhead.
Notably, our VIDEAS-WM consistently occupies this optimal region.
Despite smaller parameter scales, VIDEAS-WM not only surpasses larger models such as Gemma4-31B in embodied action semantic reasoning but also exhibits significantly reduced inference latency.

\section{Text-based Environment}

\begin{figure}[p]
    \centering

    \includegraphics[width=0.93\linewidth]{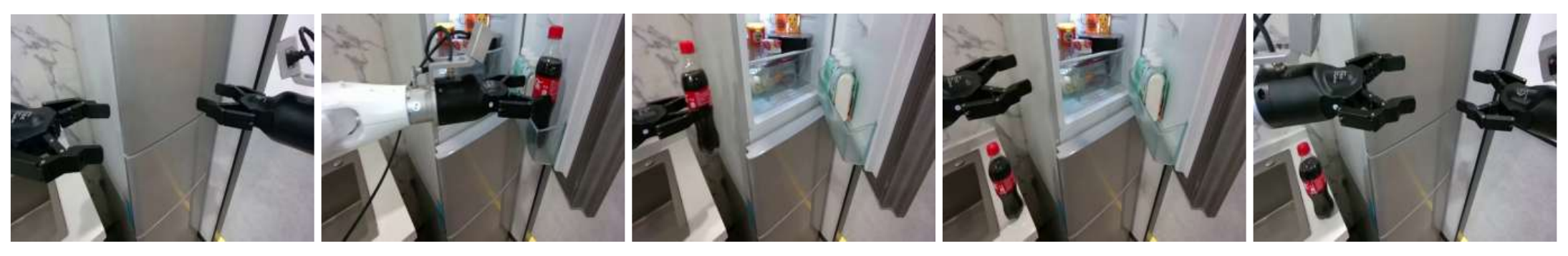}
    \vspace{1mm}

    \begin{minipage}{0.93\linewidth}
        \hrule height 0.8pt
        \vspace{1mm}

        \textbf{Task Name:} Open the fridge to get food. \\
        \textbf{Action Trajectory:}
        (1) Open the refrigerator door with the left arm.
        $\rightarrow$
        (2) Pick up the cola from the fridge with the left arm.
        $\rightarrow$
        (3) Place the cola on the table.
        $\rightarrow$
        (4) Close the refrigerator door with the right arm.

        \vspace{1mm}
        \hrule
        \vspace{1mm}

        \textbf{Text-Based Embodied Environment (Abbreviated JSON Structure)}

\begin{lstlisting}[
    basicstyle=\ttfamily\fontsize{7.2pt}{7.5pt}\selectfont,
    breaklines=true,
    frame=none,
    xleftmargin=0pt,
    aboveskip=1mm,
    belowskip=1mm,
    keepspaces=true
]
{
    "rooms": {
        "kitchen": {
            "vertices": [[0, 0], [4, 0], [4, 5], [0, 5]],
            "floor": "white ceramic tile, glossy",
            "wall": "light beige drywall, smooth"
        }
    },
    "...": "...",
    "agents": {
        "service_robot": {
            "description": "A humanoid service robot with two arms for manipulation",
            "size": 1.2,
            "gripper_contact_range": 1.2,
            "located_room": "kitchen",
            "position": [3.62, 3.87, 0.6]
        }
    },
    "spatial_relation": {
        "cola_bottle_refrigerator_1": {
            "source": "cola_bottle",
            "type": "in",
            "target": "refrigerator",
            "parameter": null
        },
        "refrigerator_north_1": {
            "source": "refrigerator",
            "type": "edge",
            "target": "north",
            "parameter": null
        },
        "...": "..."
    },
    "non_spatial_relation": {},
    "objects": {
        "refrigerator": {
            "description": "A stainless steel household refrigerator with a single door",
            "size": [0.7, 0.8, 1.8],
            "located_room": "kitchen",
            "open_close_state": "closed",
            "content": ["cola_bottle"],
            "power_state": "on",
            "position": [3.62, 4.56, 0.9],
            "direction": 180
        },
        "cola_bottle": {
            "description": "A plastic bottle of cola with a screw cap",
            "size": [0.08, 0.08, 0.2],
            "located_room": "kitchen",
            "content": [["cola"]],
            "position": [3.5, 4.5, 1.0],
            "direction": 0
        },
        "kitchen_table": {
            "description": "A rectangular wooden table with a smooth surface",
            "size": [1.0, 0.8, 0.75],
            "located_room": "kitchen",
            "position": [3.56, 3.0, 0.37],
            "direction": 270
        },
        "...": "..."
    }
}
\end{lstlisting}

        \vspace{-1mm}
        \hrule height 0.8pt
    \end{minipage}

    \vspace{-1mm}
    \caption{An illustrative example of the text-based environment. The complete JSON structure encompasses comprehensive fine-grained properties, which are abbreviated here for brevity.}
    \label{fig:env-example}
\end{figure}

We provide a concrete example to illustrate the structure of text-based simulated environments, as shown in Figure \ref{fig:env-example}.
It presents a typical kitchen scenario where the agent receives instructions to interact with household appliances and retrieve target objects.
The environment state is formalized into a hierarchical JSON format, capturing room layouts, physical entities, spatial relations, and non-spatial relations.
This structured context provides the VLMs with a fully observable and deterministic physical world to determine action preconditions and track state transitions.

\section{Prompt}
\label{app:prompt}

\begin{figure*}[!htbp]
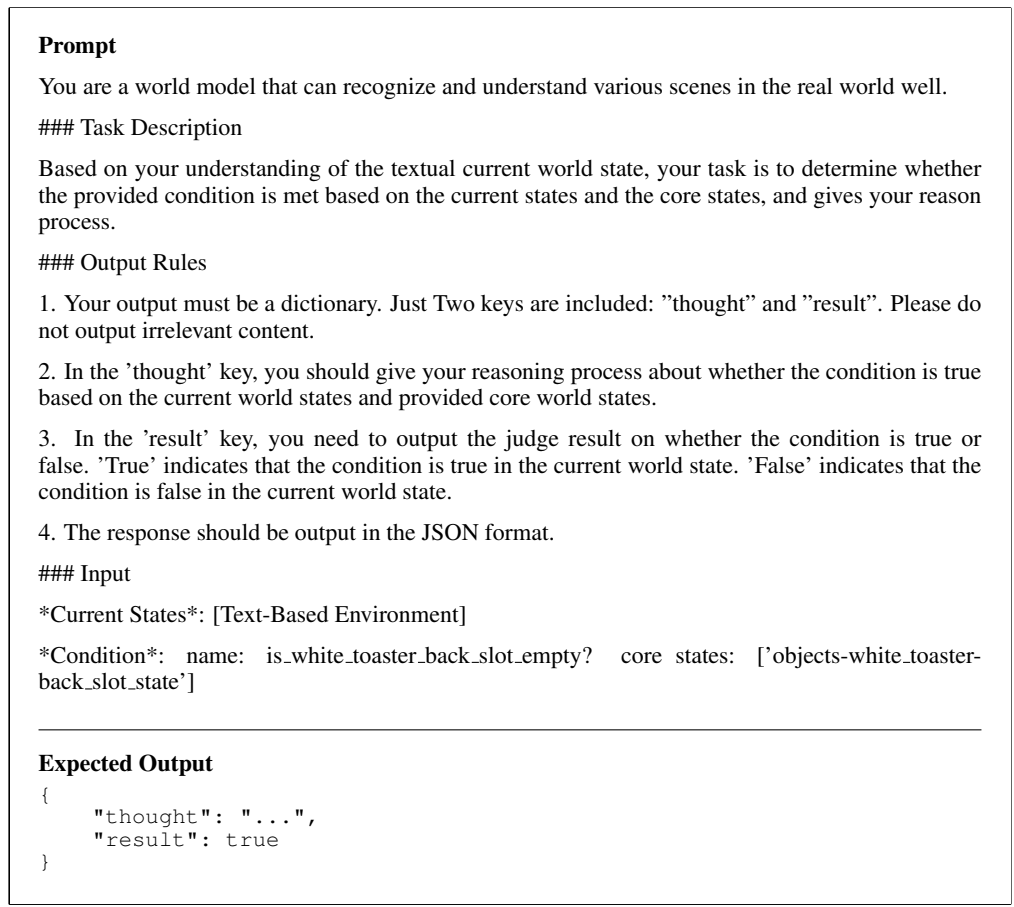

\centering
\begin{tcolorbox}[colback=white, colframe=black, boxrule=0.5pt, width=0.95\textwidth, arc=0pt, left=8pt, right=8pt, top=8pt, bottom=8pt]
\small
\setlength{\parindent}{0pt}
\setlength{\parskip}{0.6em}
\textbf{Prompt}

You are a world model that can recognize and understand various scenes in the real world well.

\#\#\# Task Description

Based on your understanding of the textual current world state, your task is to determine whether the provided condition is met based on the current states and the core states, and gives your reason process.

\#\#\# Output Rules

1. Your output must be a dictionary. Just Two keys are included: "thought" and "result". Please do not output irrelevant content.

2. In the 'thought' key, you should give your reasoning process about whether the condition is true based on the current world states and provided core world states.

3. In the 'result' key, you need to output the judge result on whether the condition is true or false. 'True' indicates that the condition is true in the current world state. 'False' indicates that the condition is false in the current world state.

4. The response should be output in the JSON format.

\#\#\# Input

*Current States*:
[Text-Based Environment]

*Condition*:
name: is\_white\_toaster\_back\_slot\_empty?
core states: ['objects-white\_toaster-back\_slot\_state']

\noindent\rule{\linewidth}{0.4pt}

\textbf{Expected Output}

\begin{lstlisting}[
    basicstyle=\ttfamily\fontsize{8.4pt}{8.4pt}\selectfont,
    breaklines=true,
    breakatwhitespace=false,
    columns=fullflexible,
    keepspaces=true,
    frame=none,
    xleftmargin=0pt,
    xrightmargin=0pt,
    aboveskip=1mm,
    belowskip=0mm,
    showstringspaces=false
]
{
    "thought": "...",
    "result": true
}
\end{lstlisting}

\end{tcolorbox}
\caption{The prompt structure and expected output for the \textit{\textbf{decide}} stage.}
\label{fig:prompt_decide}
\end{figure*}

\begin{figure*}[!htbp]
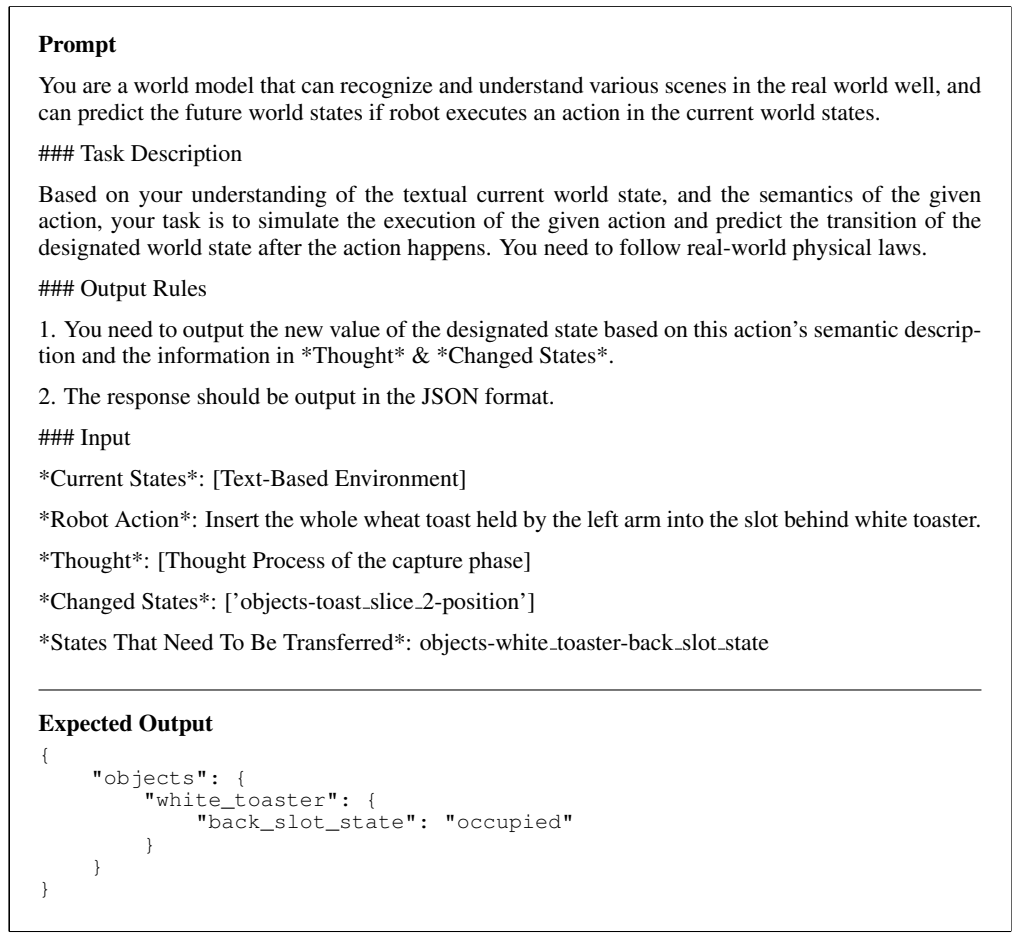

\centering
\begin{tcolorbox}[colback=white, colframe=black, boxrule=0.5pt, width=0.95\textwidth, arc=0pt, left=8pt, right=8pt, top=8pt, bottom=8pt]
\small
\setlength{\parindent}{0pt}
\setlength{\parskip}{0.6em}
\textbf{Prompt}

You are a world model that can recognize and understand various scenes in the real world well, and can predict the future world states if robot executes an action in the current world states.

\#\#\# Task Description

Based on your understanding of the textual current world state, and the semantics of the given action, your task is to simulate the execution of the given action and predict the transition of the designated world state after the action happens. You need to follow real-world physical laws.

\#\#\# Output Rules

1. You need to output the new value of the designated state based on this action's semantic description and the information in *Thought* \& *Changed States*.

2. The response should be output in the JSON format.

\#\#\# Input

*Current States*:
[Text-Based Environment]

*Robot Action*:
Insert the whole wheat toast held by the left arm into the slot behind white toaster.

*Thought*:
[Thought Process of the capture phase]

*Changed States*:
['objects-toast\_slice\_2-position']

*States That Need To Be Transferred*:
objects-white\_toaster-back\_slot\_state

\noindent\rule{\linewidth}{0.4pt}

\textbf{Expected Output}

\begin{lstlisting}[
    basicstyle=\ttfamily\fontsize{8.2pt}{8.4pt}\selectfont,
    breaklines=true,
    breakatwhitespace=false,
    columns=fullflexible,
    keepspaces=true,
    frame=none,
    xleftmargin=0pt,
    xrightmargin=0pt,
    aboveskip=1mm,
    belowskip=0mm,
    showstringspaces=false
]
{
    "objects": {
        "white_toaster": {
            "back_slot_state": "occupied"
        }
    }
}
\end{lstlisting}

\end{tcolorbox}
\caption{The prompt structure and expected output for the \textit{\textbf{transfer}} stage.}
\label{fig:prompt_transfer}
\end{figure*}

This section details the prompts used in the trajectory simulation mechanism.
The main text states that this mechanism models the execution of atomic actions through three phases to provide a concise high-level overview.
The practical implementation operates through four distinct computational stages.
We decouple the final transfer phase into a capture stage and a subsequent transfer stage.
This decoupling reduces the reasoning burden on VLMs and ensures higher fidelity in predicting complex environmental dynamics.
Using the Toast Bread episode as an example, we present the prompts and ideal VLM responses corresponding to the four stages of a specific action (Figure \ref{fig:prompt_consider} - \ref{fig:prompt_transfer}).

\begin{figure}[p]
\centering
\begin{tcolorbox}[colback=white, colframe=black, boxrule=0.5pt, width=0.95\textwidth, arc=0pt, left=8pt, right=8pt, top=8pt, bottom=8pt]
\small
\setlength{\parindent}{0pt}
\setlength{\parskip}{0.6em}
\textbf{Prompt}

You are a world model that can recognize and understand various scenes in the real world well, and can determine whether the action could be executed in the current scene.

\#\#\# Task Description

Based on your understanding of the textual current world states, the semantics of the given robot action, and the environmental observations before and after action execution (start frame and end frame), your task is to think the preconditions of the robot action. And you need to give your reason process.

\#\#\# Output Rules

1. Your output must be a dictionary. Just three keys are included: 'thought', 'corestates', and 'corestates\_successtag'. Please do not output irrelevant content.

2. In the 'thought' key, you should first summarize the conditions that need to be met and the corresponding boolean value for the action to be executed successfully. Then, identify which states in the textual current world states are crucial for influencing each condition. Boolean value is true, indicating that these conditions should be met for the action execution; Boolean value is false, indicating that these conditions should not be met for the action execution.

3. In the 'corestates' key, the value is a dictionary. The dictionary includes all preconditions that affect the execution of the action. The keys of the dictionary should be expressed as complete question sentences, representing each precondition. The corresponding values should be the core states from the textual current world states that are necessary to check each precondition. The state names should be represented as A-B-C. Keys from different levels are connected with hyphen. Each condition corresponds to several states in a list.

4. In the 'corestates\_successtag' key, it shows the boolean value that each precondition in the 'corestates' dictionary should return for the action to be executed successfully. Ensure that information in 'corestates\_successtag' should be consistent with the meaning in 'corestates'.

5. You don't need to consider whether the robot's grippers are free.

6. The response should be output in the JSON format.

\#\#\# Input

*Current States*:
[Text-Based Environment]

*Robot Action*:
Insert the whole wheat toast held by the left arm into the slot behind white toaster.

*Environmental Observations*:
start frame: [Start Frame]
end frame: [End Frame]

\noindent\rule{\linewidth}{0.4pt}

\textbf{Expected Output}
\vspace{0.2em}

\begin{lstlisting}[
    basicstyle=\ttfamily\fontsize{8.4pt}{8.4pt}\selectfont,
    breaklines=true,
    breakatwhitespace=false,
    columns=fullflexible,
    keepspaces=true,
    frame=none,
    xleftmargin=0pt,
    xrightmargin=0pt,
    aboveskip=1mm,
    belowskip=0mm,
    showstringspaces=false
]
{
    "thought": "...",
    "corestates_successtag": {
        "whether_robot_hold_toast_slice_2?": true,
        "is_white_toaster_back_slot_empty?": true,
        "is_white_toaster_inside_robot_gripper_contact?": true
    },
    "corestates": {
        "whether_robot_hold_toast_slice_2?": [
            "non_spatial_relation-toasting_robot_toast_slice_2_19704749-type"
        ],
        "is_white_toaster_back_slot_empty?": [
            "objects-white_toaster-back_slot_state"
        ],
        "is_white_toaster_inside_robot_gripper_contact?": [
            "agents-toasting_robot-position",
            "objects-white_toaster-position",
            "agents-toasting_robot-gripper_contact_range"
        ]
    }
}
\end{lstlisting}
\end{tcolorbox}
\caption{The prompt structure and expected output for the \textit{\textbf{consider}} stage.}
\label{fig:prompt_consider}
\end{figure}

\begin{figure*}[!htbp]
\centering
\begin{tcolorbox}[colback=white, colframe=black, boxrule=0.5pt, width=0.95\textwidth, arc=0pt, left=8pt, right=8pt, top=8pt, bottom=8pt]
\small
\setlength{\parindent}{0pt}
\setlength{\parskip}{0.6em}
\textbf{Prompt}

You are a world model that can recognize and understand various scenes in the real world well, and can predict the future world states if robot executes an action in the current world states.

\#\#\# Task Description

Based on your understanding of the textual current world states, the semantics of the given robot action, and the environmental observations before and after action execution (start frame and end frame), your task is to think all world states that need to be transferred after the action is executed. Based on the start frame and end frame, you need to compare the environment changes before and after the action execution to enhance reasoning. And please give your reason process.

\#\#\# Output Rules

1. Your output must be a dictionary. Just Two keys are included: "thought" and "states\_transferred". Please do not output irrelevant content.

2. For "thought" key, please output your thought and reason process about the current situation and which states needed to be transferred.

3. For "states\_transferred" key, its value is a dictionary which includes three keys: "insert", "delete" and "update". They respectively represent the relations to be added, the relations to be deleted and the states to be updated.

4. For "insert" key, its value is a dictionary including two keys: "spatial\_relation" and "non\_spatial\_relation". Their value is a nested list where each sub-list is a string list of length 3, respectively the source party, the relation type and the object party. 

5. For "delete" key, its value is a list which indicates the relations to be deleted. Each relation should be represented as A-B. Keys from different levels are connected with hyphens.

6. For "update" key, its value is a list which indicates the states to be updated. Each state should be represented as A-B-C. Keys from different levels are connected with hyphens.

7. The response should be output in the JSON format.

\#\#\# Input

*Current States*:
[Text-Based Environment]

*Robot Action*:
Insert the whole wheat toast held by the left arm into the slot behind white toaster.

*Environmental Observations*:
start frame: [Start Frame]
end frame: [End Frame]

\noindent\rule{\linewidth}{0.4pt}

\textbf{Expected Output}

\begin{lstlisting}[
    basicstyle=\ttfamily\fontsize{8.4pt}{8.4pt}\selectfont,
    breaklines=true,
    breakatwhitespace=false,
    columns=fullflexible,
    keepspaces=true,
    frame=none,
    xleftmargin=0pt,
    xrightmargin=0pt,
    aboveskip=1mm,
    belowskip=0mm,
    showstringspaces=false
]
{
    "thought": "...",
    "states_transferred": {
        "insert": {
            "spatial_relation": [
                [
                    "toast_slice_2",
                    "in",
                    "white_toaster"
                ]
            ],
            "non_spatial_relation": []
        },
        "delete": [
            "non_spatial_relation-toasting_robot_toast_slice_2_19704749"
        ],
        "update": [
            "objects-toast_slice_2-position",
            "objects-white_toaster-back_slot_state"
        ]
    }
}
\end{lstlisting}

\end{tcolorbox}
\caption{The prompt structure and expected output for the \textit{\textbf{capture}} stage.}
\label{fig:prompt_capture}
\end{figure*}

\section{Negative Trajectory Case}

\begin{figure*}[htbp]
    \centering
    
    \includegraphics[width=0.95\linewidth]{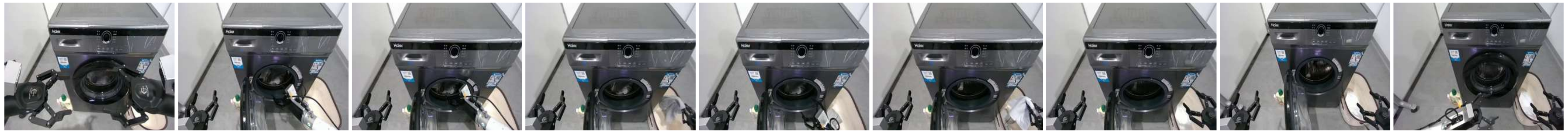}
    \vspace{0.3cm}
    
    \begin{minipage}{0.95\linewidth}
        \hrule height 1pt
            \vspace{0.2cm}
            \textbf{Task Name:} Remove clothes from the washing machine \\
            \textbf{Action Trajectory:} (1) Open the washing machine door with both arms. $\rightarrow$ (2) Move forward to approach the washing machine. $\rightarrow$ (3) Retrieve shorts from the washing machine with the right arm. $\rightarrow$ (4) Place the shorts into the laundry basket. $\rightarrow$ (5) Retrieve a short sleeve from the washing machine. $\rightarrow$ (6) Place the short sleeve into the laundry basket. $\rightarrow$ (7) Step back a bit away from the washing machine. $\rightarrow$ (8) Close the washing machine door. \\
        \hrule
            \vspace{0.15cm}
            \textbf{Dependency Records (Abbreviated JSON Structure)}
\begin{lstlisting}[basicstyle=\ttfamily\fontsize{8.2pt}{8.8pt}\selectfont,, breaklines=true, frame=none, xleftmargin=0.2cm, keepspaces=true]
{
    "778774": {
        "task_name": "Remove clothes from the washing machine",
        "actions": [
            {"index": 0, "name": "0_Open", "skill": "Open", "text": "Open the washing machine door with both arms."},
            {"index": 1, "name": "1_Move", "skill": "Move", "text": "Move forward to approach the washing machine."},
            {"index": 2, "name": "2_Pick", "skill": "Pick", "text": "Retrieve a shorts from the washing machine with the right arm."},
            ...
        ],
        "dependencies": [
            {
                "action_index": 2,
                "action_name": "2_Pick",
                "depends_on": [0],
                "dependency_type": "object_state",
                "precondition_nl": "washing machine door is open",
                "mutation_viable": true
            },
            {
                "action_index": 3,
                "action_name": "3_Place",
                "depends_on": [2],
                "dependency_type": "hold_object",
                "precondition_nl": "robot is holding a shorts",
                "mutation_viable": true
            },
            {
                "action_index": 4,
                "action_name": "4_Pick",
                "depends_on": [0],
                "dependency_type": "object_state",
                "precondition_nl": "washing machine door is open",
                "mutation_viable": true
            },
            "...": "..."
        ]
    }
}
\end{lstlisting}
            \vspace{-0.2cm}
            \hrule
            \vspace{0.2cm}
            \textbf{Mutated Negative Trajectory:} (1) Move forward to approach the washing machine. $\rightarrow$ (2) Retrieve a shorts from the washing machine with the right arm. $\rightarrow$ (3) Place the shorts into the laundry basket. $\rightarrow$ (4) Retrieve a short sleeve from the washing machine. $\rightarrow$ (5) Place the short sleeve into the laundry basket. $\rightarrow$ (6) Step back a bit away from the washing machine. $\rightarrow$ (7) Close the washing machine door.
        \vspace{0.2cm}
        \hrule height 1pt
    \end{minipage}
    
    \caption{An illustrative example of negative trajectories. By removing the initial \textit{Open} action, the subsequent \textit{Pick} action is deliberately designed to fail due to unmet physical preconditions.}
    \label{fig:negative-case}
\end{figure*}

Figure \ref{fig:negative-case} illustrates the synthesis process of a negative trajectory, exemplified by a representative episode of \textit{retrieving clothes from a washing machine}.

\section{Case Study}

In Figure \ref{fig:case-study-part1}, we present a case study illustrating how VIDEAS-WM-9B simulates a predefined robot policy for the BoilWater task from the BTSIMBENCH benchmark.

\begin{figure*}[htbp]
\centering
\begin{minipage}{0.95\linewidth}
\hrule height 1pt
\vspace{0.2cm}
\textbf{Task Name:} BoilWater \\
\textbf{Task Description:} The robot task is to boil water. The behavior logic of a robot should be as follows. The kettle is initially inside a cabinet and is filled with water. The stove and cabinet are all located in the kitchen. The robot needs to move the kettle from the cabinet to the stove, heat the water, and wait for that the water in the kettle is boiled. The goal is to get boiled water.
\vspace{0.01cm}
\hrule height 1pt
\vspace{0.1cm}
\renewcommand{\arraystretch}{1.2}
\begin{tabular}{@{} p{0.35\linewidth} | p{0.61\linewidth} @{}}
\textbf{Action Sequence} & \textbf{Predicted Preconditions \& Effects (JSON Structure)} \\
\hline
(1) \textbf{Open\_cabinet}: The robot grips the cabinet door handle, pulls it open smoothly, ensuring a safe and controlled motion. &
\begin{minipage}[t]{\linewidth}
Preconditions:
\begin{lstlisting}[basicstyle=\ttfamily\scriptsize, breaklines=true, frame=none, keepspaces=true, aboveskip=0.1cm, belowskip=0.15cm]
{
  "is_cabinet_closed?": true,
  "is_cabinet_inside_robot_gripper_contact?": true
}
\end{lstlisting}
Effects:
\begin{lstlisting}[basicstyle=\ttfamily\scriptsize, breaklines=true, frame=none, keepspaces=true, aboveskip=0.1cm, belowskip=0.1cm]
{
  "states_transferred": {
    "insert": {
      "spatial_relation": [], 
      "non_spatial_relation": []
    },
    "delete": [],
    "update": [
      "objects-cabinet-open_close_state"
    ]
  }
}
\end{lstlisting}
\end{minipage} \\
\hline
(2) \textbf{Take\_out\_kettle}: The robot extends its arm, grasps the kettle, and takes it out from the cabinet. &
\begin{minipage}[t]{\linewidth}
Preconditions:
\begin{lstlisting}[basicstyle=\ttfamily\scriptsize, breaklines=true, frame=none, keepspaces=true, aboveskip=0.1cm, belowskip=0.15cm]
{
  "is_cabinet_open?": true,
  "is_kettle_in_cabinet?": true,
  "is_kettle_inside_robot_gripper_contact?": true
}
\end{lstlisting}
Effects:
\begin{lstlisting}[basicstyle=\ttfamily\scriptsize, breaklines=true, frame=none, keepspaces=true, aboveskip=0.1cm, belowskip=0.1cm]
{
  "states_transferred": {
    "insert": {
      "spatial_relation": [],
      "non_spatial_relation": [
        ["kitchen_robot", "hold", "kettle"]
      ]
    },
    "delete": [
      "spatial_relation-kettle_cabinet_1"
    ],
    "update": [
      "objects-kettle-position"
    ]
  }
}
\end{lstlisting}
\end{minipage} \\
\hline
(3) \textbf{Move\_to\_stove}: The robot moves to the location of the stove. &
\begin{minipage}[t]{\linewidth}
Preconditions:
\begin{lstlisting}[basicstyle=\ttfamily\scriptsize, breaklines=true, frame=none, keepspaces=true, aboveskip=0.1cm, belowskip=0.15cm]
{}
\end{lstlisting}
Effects:
\begin{lstlisting}[basicstyle=\ttfamily\scriptsize, breaklines=true, frame=none, keepspaces=true, aboveskip=0.1cm, belowskip=0.1cm]
{
  "states_transferred": {
    "insert": {
      "spatial_relation": [], 
      "non_spatial_relation": []
    },
    "delete": [],
    "update": [
      "agents-kitchen_robot-position",
      "objects-kettle-position"
    ]
  }
}
\end{lstlisting}
\end{minipage} \\
\end{tabular}
\vspace{0.15cm}
\hrule height 1pt
\end{minipage}
\caption{Case study illustrating the simulation of a robot policy from the BTSIMBENCH benchmark using VIDEAS-WM-9B. For each action within the sequence, the predicted preconditions and effects are formalized in JSON format.}
\label{fig:case-study-part1}
\end{figure*}

\begin{figure*}[htbp]
\ContinuedFloat
\centering
\begin{minipage}{0.95\linewidth}
\hrule height 1pt
\vspace{0.1cm}
\renewcommand{\arraystretch}{1.2}
\begin{tabular}{@{} p{0.35\linewidth} | p{0.61\linewidth} @{}}
\textbf{Action Sequence} & \textbf{Predicted Preconditions \& Effects (JSON Structure)} \\
\hline
(4) \textbf{Place\_kettle\_on\_stove}: The robot places the kettle on the stove smoothly. &
\begin{minipage}[t]{\linewidth}
Preconditions:
\begin{lstlisting}[basicstyle=\ttfamily\scriptsize, breaklines=true, frame=none, keepspaces=true, aboveskip=0.1cm, belowskip=0.15cm]
{
  "is_stove_inside_robot_gripper_contact?": true,
  "whether_robot_hold_kettle?": true
}
\end{lstlisting}
Effects:
\begin{lstlisting}[basicstyle=\ttfamily\scriptsize, breaklines=true, frame=none, keepspaces=true, aboveskip=0.1cm, belowskip=0.1cm]
{
  "states_transferred": {
    "insert": {
      "spatial_relation": [
        ["kettle", "on_top_of", "stove"]
      ],
      "non_spatial_relation": []
    },
    "delete": [
      "non_spatial_relation-kitchen_robot_kettle_c7af0617"
    ],
    "update": [
      "objects-kettle-position"
    ]
  }
}
\end{lstlisting}
\end{minipage} \\
\hline
(5) \textbf{Turn\_on\_stove}: The robot turns on the stove switch to start heating. &
\begin{minipage}[t]{\linewidth}
Preconditions:
\begin{lstlisting}[basicstyle=\ttfamily\scriptsize, breaklines=true, frame=none, keepspaces=true, aboveskip=0.1cm, belowskip=0.15cm]
{
  "is_stove_inside_robot_gripper_contact?": true,
  "is_stove_off?": true
}
\end{lstlisting}
Effects:
\begin{lstlisting}[basicstyle=\ttfamily\scriptsize, breaklines=true, frame=none, keepspaces=true, aboveskip=0.1cm, belowskip=0.1cm]
{
  "states_transferred": {
    "insert": {
      "spatial_relation": [], 
      "non_spatial_relation": []
    },
    "delete": [],
    "update": [
      "objects-stove-power_state"
    ]
  }
}
\end{lstlisting}
\end{minipage} \\
\hline
(6) \textbf{Wait\_water\_boiled}: The robot waits for the stove to heat the kettle until the water in the kettle reaches boiling point. &
\begin{minipage}[t]{\linewidth}
Preconditions:
\begin{lstlisting}[basicstyle=\ttfamily\scriptsize, breaklines=true, frame=none, keepspaces=true, aboveskip=0.1cm, belowskip=0.15cm]
{
  "is_kettle_on_stove?": true,
  "kettle_has_water?": true,
  "kettle_is_closed?": true,
  "kettle_is_not_boiled?": true,
  "stove_is_on?": true
}
\end{lstlisting}
Effects:
\begin{lstlisting}[basicstyle=\ttfamily\scriptsize, breaklines=true, frame=none, keepspaces=true, aboveskip=0.1cm, belowskip=0.1cm]
{
  "states_transferred": {
    "insert": {
      "spatial_relation": [], 
      "non_spatial_relation": []
    },
    "delete": [],
    "update": [
      "objects-kettle-temperature",
      "objects-kettle-boiling_state"
    ]
  }
}
\end{lstlisting}
\end{minipage} \\
\end{tabular}
\vspace{0.15cm}
\hrule height 1pt
\end{minipage}
\caption{Case study illustrating the simulation of a robot policy from the BTSIMBENCH benchmark using VIDEAS-WM-9B (Continued).}
\label{fig:case-study-part2}
\end{figure*}

\end{document}